%% file: acl_latex.tex
\pdfoutput=1

\documentclass[11pt]{article}

\usepackage{acl}

\usepackage{xcolor}
\usepackage{adjustbox}
\usepackage[most]{tcolorbox}
\usepackage{multirow}
\usepackage{booktabs}
\usepackage{colortbl}
\usepackage{times}
\usepackage{inconsolata}
\usepackage{latexsym}
\usepackage{graphicx}
\usepackage{subcaption} 
\usepackage{algorithm,algpseudocode}
\usepackage{
  amsfonts, nicefrac, amsmath, bm,
  soul, graphicx, multirow, wrapfig, lipsum, booktabs, caption, subcaption, stfloats, listings,
  longtable, paralist, cleveref, amssymb, pifont
}

\usepackage{authblk}
\usepackage{enumitem}
\usepackage[T1]{fontenc}
\usepackage[utf8]{inputenc}
\usepackage{microtype}
\usepackage{mparhack}
\usepackage{hyperref}

\newcommand{\gpt}{\texttt{gpt-4o}}
\newcommand{\gptmini}{\texttt{gpt-4o-mini}}
\newcommand{\llamabig}{\texttt{Llama-3.1-70b-Instruct}}
\newcommand{\llamasmall}{\texttt{Llama-3.1-8b-Instruct}}
\newcommand{\rr}{\texttt{RequestRepair}}
\newcommand{\rep}{\texttt{Repair}}

\definecolor{MutedOrange}{RGB}{255, 235, 200}
\definecolor{MutedBlue}{RGB}{200, 200, 255}

\newtcolorbox[list inside=prompt,auto counter,number within=section]{prompt}[1][]{
    colbacktitle=black!60,
    fonttitle=\small,
    coltitle=white,
    fontupper=\footnotesize,
    boxsep=4pt,
    left=0pt,
    right=0pt,
    top=0pt,
    bottom=0pt,
    boxrule=1pt,
    #1,
}

\newcommand{\umcp}{University of Maryland, College Park}
\newcommand{\orau}{Oak Ridge Associated Universities}
\newcommand{\arl}{Army Research Lab}

\crefname{section}{\S}{\S\S}
\title{Understanding Common Ground Misalignment in Goal-Oriented Dialog: A Case-Study with Ubuntu Chat Logs}

\author[1]{\textbf{Rupak Sarkar}} 
\author[1]{\textbf{Neha Srikanth}} 
\author[2]{\textbf{Taylor Pellegrin}} 
\author[1]{\\ \textbf{Rachel Rudinger}}
\author[3]{\textbf{Claire Bonial}}
\author[1]{\textbf{Philip Resnik}}
\affil[1]{\umcp}
\affil[2]{\orau}
\affil[3]{\arl}
\affil[ ]{\texttt{\{rupak,nehasrik\}@umd.edu}}

\begin{document}
\maketitle

\begin{abstract}
    \input{sections/00-abstract}
\end{abstract}

\section{Introduction}
\label{sec:intro}
\input{sections/10-intro}

\section{Background}
\label{sec:motivation}
\input{sections/20-motivation}

\section{Conversational Friction}
\label{sec:approach}
\input{sections/30-approach}

\section{Analysis of Grounding in Ubuntu-CG}
\label{sec:human}
\input{sections/40-human}

\section{Can LLMs Identify Conversational Friction?}
\label{sec:llm}
\input{sections/50-llm}

\section{Error Analysis}
\label{sec:error}
\input{sections/55-error_analysis}

\input{sections/65-related_work}

\input{sections/70-conclusion}

\section*{Limitations}
\label{sec:limitations}
\input{sections/75-limitations}

\bibliography{anthology, custom}

\clearpage
\appendix
\input{sections/80-appendix}

\clearpage

\end{document}

%% file: sections/00-abstract.tex
While it is commonly accepted that maintaining common ground plays a role in conversational success, little prior research exists connecting conversational grounding to success in task-oriented conversations. 
We study failures of grounding in the Ubuntu~IRC dataset, where participants use text-only communication to resolve technical issues. 
We find that disruptions in conversational flow often stem from a misalignment in common ground, driven by a divergence in beliefs and assumptions held by participants. 
These disruptions, which we call conversational friction, significantly correlate with task success. 
While LLMs can identify overt cases of conversational friction, they struggle with subtler and more context-dependent instances that require pragmatic or domain-specific reasoning.

%% file: sections/10-intro.tex
Effective communication between humans in conversation hinges on a set of facts and beliefs relevant to the conversation, or the \textit{conversational common ground}~\cite{stalnaker1978,Stalnaker2002,clark1991grounding}, that is shared between participants. 
They must collaboratively maintain and update this common ground for the conversation to progress successfully.
This dynamic, ongoing management is essential: a misalignment or misunderstanding can disrupt the communicative flow, potentially leading to confusion or conflict.

\begin{figure}[!h]
\centering
\includegraphics[scale=0.78]{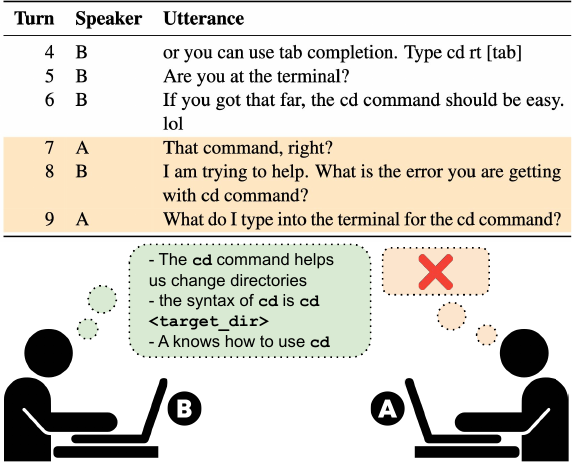}
\caption{An annotated instance of conversational friction. 
Though it is challenging to access propositions in a speakers' 
   perception of common ground, certain propositions in B's version of common ground are revealed (green thought bubble) when there is a misalignment between the two participants. B assumes A knows about the \texttt{cd} command, which is proven false by A in Turn 9.}
\label{fig:teaser}
\end{figure}

Typically, much of this maintenance is implicit: listeners acknowledge their understanding through verbal and non-verbal cues, making research on common ground and its role in conversational success challenging.
When participants successfully complete a goal-oriented conversation without visible disruption or misunderstanding, it is unclear what information constitutes their common ground. 
Many studies sidestep this by constraining the conversational setting to physically grounded tasks, such as building objects in Minecraft-like worlds~\cite{narayan-chen-etal-2019-collaborative, bara-etal-2021-mindcraft}, providing environments where researchers can infer participants' common ground through their actions.

We address this challenge in a different way---by focusing on \emph{miscommunications} as a window into the shared beliefs of conversational participants. 
Consider the conversation in Figure~\ref{fig:teaser}.
At the outset, the common ground includes beliefs such as \emph{``A is an Ubuntu user''} and \emph{``A is accessing a Linux terminal''}, etc.
Following Turn~4, B believes that \emph{``the syntax of \texttt{cd} is \texttt{cd $\langle \mbox{target\_dir}\rangle$}}'' is now part of the conversational common ground.
Turn~9 reveals that this assumption was incorrect through an \emph{observable} interruption precluding A and B from proceeding towards A's main conversational goal.\footnote{\citet{grosz1986attention} would distinguish this goal as the \emph{discourse purpose}.}

We use the term \textbf{conversational friction} to describe such an instance of disruption in communicative flow, caused by a misalignment in speaker beliefs about what is present in the common ground.\footnote{Hereafter we use the terms ``friction'' and ``conversational friction'' interchangeably.}
Frictions reveal the importance of maintaining common ground, as they require re-negotiation~\cite{clark1986referring} of content: instead of making progress, participants need a ``conversational detour'' to align their interpretations of previously shared content. 

This work explores two key questions. First, (\textbf{RQ1}) to what extent is achieving a participant's goal---or \textit{success}---associated with the presence or absence of conversational friction? 
And (\textbf{RQ2}), can large language models (LLMs) identify and explain sources of friction in human conversations? 
We seek to shed light on the relationship between conversational friction, which serves as evidence of a misalignment in common ground, and the success of participants in achieving a shared goal. 

To achieve this, we annotate real-world conversations involving Ubuntu users attempting to fix an issue or bug.\footnote{We will release code and data upon publication.}
We annotate 200 conversations from the Ubuntu Dialog Corpus~\cite{kummerfeld-etal-2019-large}, a corpus of conversations among users solving issues when using the Ubuntu operating system.\footnote{Ubuntu (\url{https://ubuntu.com/desktop}) is one of the most popular free and open-source Linux-based operating systems in the world.}
Each conversation is annotated for the presence of conversational friction (supporting RQ2) and the degree of task success (supporting RQ1) (\S\ref{sec:friction-annotation}) to analyze the importance of maintaining common ground (\S\ref{sec:human}).
Then, we explore the ability of LLMs to identify friction and compare their explanations with human explanations (\S\ref{sec:llm}). 

Not only are LLMs increasingly relied upon as conversational partners~\cite{minaee2024largelanguagemodelssurvey}, they are also used as mediators~\cite{tan2024robotsmiddleevaluatingllms} or to generate conversational summaries~\cite{ramprasad2024analyzingllmbehaviordialogue}.
As such, it is important to know if they track the common ground, an essential component of smooth communication.
Our analyses of friction and repair reveal that \textbf{friction often arises from misalignment in common ground}, particularly when participants hold diverging assumptions about the task or possess varying levels of domain expertise. 
Furthermore, we find that while models are able to detect overt signals of friction, \textbf{they struggle to identify subtler and more context-dependent instances of misalignment} that require deeper pragmatic or domain-specific reasoning\footnote{Code and data can be found in \url{https://github.com/styx97/cg-misalignment}}.

\begin{figure*}[htb]
    \centering
    \includegraphics[width=1.0\textwidth]{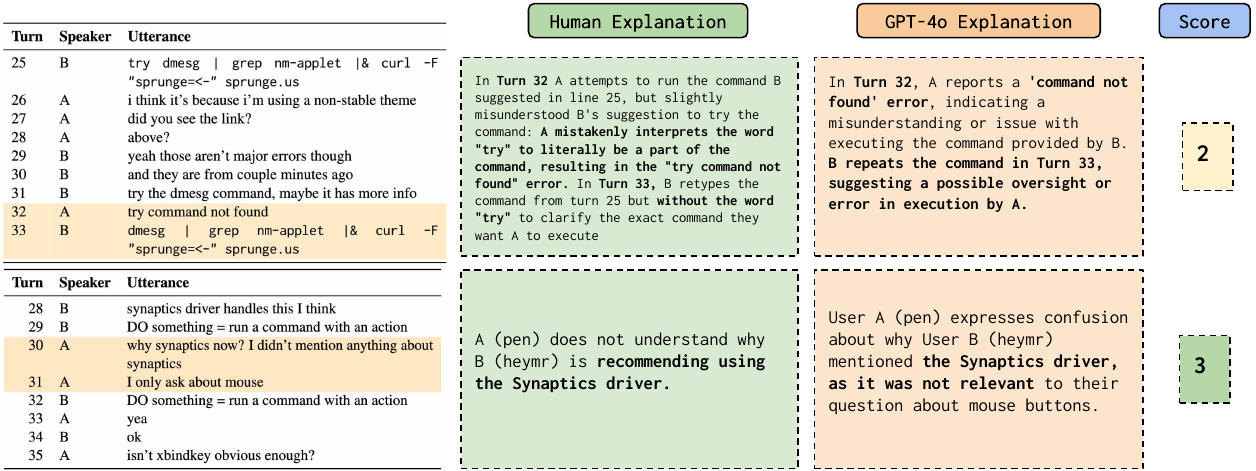}
    \caption{Comparing GPT4o and human explanations for the cause of friction. GPT4o explanations align with humans when friction is explicit (row 2). In a more implicit case of friction (row 1), GPT4o fails to capture the true reason for friction---A misreading ``try'' as part of a terminal command (Turn 25), revealed in the error message ``try command not found'' in Turn 32.}
  \label{fig:gpt4-fail}
\end{figure*}

%% file: sections/20-motivation.tex
The \textit{conversational common ground} (CG) is a body of statements treated as mutual knowledge among participants~\cite{stalnaker1978}.
It guides both how speakers choose their utterances and how they want them to be interpreted~\cite{Stalnaker2002}\footnote{Even before Stalnaker, Paul Grice mentioned propositions having \textit{common ground status} in his William James lectures~\cite{Stalnaker2002}. For a thorough discussion of common ground in linguistics, see \citet{sep-common-ground-pragmatics}.}.
Subsequently, \citet{clark1991grounding}
define common ground as a collection of mutual knowledge, beliefs, and assumptions that humans maintain collaboratively through the process of \textit{grounding}.\footnote{We focus only on discourse-theoretic grounding and do not delve into symbol grounding \cite{harnad1990symbol}, as exemplified in mapping a linguistic concept to a visual scene (see \citet{cohen2024survey} for a survey of methodologies for robotic language grounding); however, we embrace the conceptual relationship between both types of grounding, as described in \citet{chandu2021grounding}.} %

In early computational work studying CG, \citet{Traum1992} proposes breaking down a conversation into Discourse Units, where humans collaboratively build common ground through speech acts such as \texttt{RequestRepair}, a speech act through which the speaker urges their conversational partner to ground an utterance.\footnote{See Table 1 of ~\citet{Traum1992} for a full list.} 

\input{tables/dataset-stats}

While it is acknowledged that maintaining CG is of some importance to conversational success~\cite{Traum1995}, there has been little empirical work that explicitly ties participant effort in maintaining it to the success of an end goal.
In this study, we look at the importance of grounding in the success of naturally-occurring goal-oriented conversations.
\textbf{Specifically, we focus on conversational friction as evidence of the loss and re-negotiation of common ground.}

Typical conversations in our dataset (e.g., in Table \ref{tab:nvidia-conversation}) involve two participants (an \texttt{asker} and a \texttt{helper}) collaboratively attempting to solve a Linux bug over a text channel. 
This consists of several steps---the \texttt{asker} must describe their issue (often with insufficient knowledge of Linux), and the \texttt{helper} must understand their goal to propose a solution.
This makes the setting well-positioned for studying friction and grounding.

\paragraph{Dataset: Ubuntu-CG.}
The Ubuntu Dialog Corpus satisfies several criteria for our study; (1) conversations are \emph{naturally} goal-oriented (e.g., resolving an error in Ubuntu), incentivizing participants to communicate effectively; (2) participants have to establish CG from scratch; (3) conversations are text-only; and (4) are multi-turn, ranging from three turns to over 100, giving users ample time to build and utilize CG. \S\ref{sec:related-work} discusses other datasets we considered. 

The Ubuntu Dialog Corpus~\cite{lowe-etal-2015-ubuntu} contains conversations scraped from the \texttt{\#Ubuntu} IRC channel, where users primarily discuss features, issues, and bugs related to the Ubuntu operating system.
This requires disentangling conversations from a single message stream.
\citet{kummerfeld-etal-2019-large} found that the disentanglement strategy originally used had a high error rate, and released a cleaner version.
We use a sample of 200 two-person conversations from this cleaner corpus, upsampling longer conversations to study diverse behavior (Table \ref{tab:conversation_statistics}). 
We refer to this subset as Ubuntu-CG (Common Ground). 

\input{tables/success}

%% file: tables/dataset-stats.tex
\begin{table}[t!]
\scriptsize
\setlength{\tabcolsep}{4pt}
\centering
\resizebox{\columnwidth}{!}{%
\begin{tabular}{lcccc}
\toprule
& \textbf{\shortstack{Kummerfeld \\ et al. (2019)}} & \textbf{\shortstack{2-person \\ conversations}} & \textbf{\shortstack{Ubuntu-CG}} & \textbf{\shortstack{Analysis \\ Subset}} \\ 
\midrule
\textbf{\#Conversations} & 496469 & 282027 & 200 & 70 \\ 
\textbf{Average Length} & 7.16 & 5.84 & 39.75 & 51.78 \\ 
\bottomrule
\end{tabular}%
}
\caption{Overview of our dataset. We use 200 dyadic conversations sampled from \citet{kummerfeld-etal-2019-large} totaling 7590 turns for friction detection, and a subset of 70 for grounding act annotation (\S\ref{sec:grounding-acts})}
\label{tab:conversation_statistics}
\end{table}

%% file: tables/success.tex
\begin{table}[t!]
\scriptsize %
\setlength{\tabcolsep}{4pt} %
\centering
\resizebox{\columnwidth}{!}{%
\begin{tabular}{lccc}
\toprule
\textbf{Success} & \textbf{\shortstack{Mean Length \\ (Std.)}} & \textbf{\shortstack{Friction \\ (\%Present)}} & \textbf{\shortstack{Mean \#Friction \\ (when Present)}} \\ 
\midrule
1 (No Progress) & 31.90 (24.99) & 57.60 (30/52) & 2.43 \\ 
\midrule
2 (Some Progress) & 43.86 (25.73) & 55.05 (49/89) & 2.06 \\ 
3 (Success) & 40.45 (28.84) & 50.84 (30/59) & 2.13 \\ 
\bottomrule
\end{tabular}%
}
\caption{An overview of Ubuntu-CG, annotated for friction and task success. Conversations where participants make \textit{some} progress towards their task contain lower occurrences of friction (Column 4).}
\label{tab:success_aggregations}
\end{table}

%% file: sections/30-approach.tex
We now focus on detecting and understanding causes of conversational friction in Ubuntu-CG. 
Users with varying levels of expertise or familiarity with Linux and English try to collaboratively fix an issue with Ubuntu over text.\footnote{Some of the users are (self-professed) non-native speakers of English.}
This setting naturally lends itself to frequent occurrence of conversational friction. 
But how often is friction resolved in subsequent grounding, and does it have a demonstrable effect on the success of a conversation? 
To answer these questions, we collect a dataset of instances of conversational friction.

\paragraph{Task.} Given a conversation consisting of $M$ turns, the task is to identify a \textbf{list of turn intervals} $\{I_1=m_{x_1}...m_{y_1}, I_2=m_{x_2}...m_{y_2}, ...\}$ exhibiting conversational friction, or instances of disruption in communicative flow caused by a misalignment in speaker beliefs \emph{about what is present in the CG}, along \textbf{with an explanation} of why each interval exhibits friction.
A strong indicator of conversational friction is when a participant asks the other participant to \textit{repair} their conversation.
However, \textit{implicit} cases of friction require identifying when a user is struggling to keep up with the conversation.
Note that not all followup questions indicate friction. 
For example, clarification questions that ask for information not assumed to be in the CG are \textit{not} cases of conversational friction. 

\subsection{Annotating Conversational Friction}\label{sec:friction-annotation}
Three computer science undergraduates familiar with Linux annotate conversations by (1) identifying turn intervals which exhibit friction along with explanations and (2) judging the success of overall conversation on a three-point scale with respect to the conversational goal. 
Annotators were paid \$18/hr to annotate 200 conversations totaling 7950 turns, taking over 80 hours to complete.
Since conversations date back to over a decade ago, they often contain antiquated terms or references that annotators were unfamiliar with. 
To mitigate this, we provide explanations generated by \texttt{gpt-4o}~\cite{openai2024gpt4o} of technical terms in dialog turns. 
For example, the model-generated elaboration in Table \ref{tab:elaboration} (Row 1) in Appendix explains that \textit{``dapper''} and \textit{``feisty''} refer to Ubuntu versions 6.06 and 7.04. 
We make these elaborations available to various models in our computational experiments as well.

\paragraph{Conversational Success.}\label{sec:task-success} In addition to friction, annotators assess how successful participants were in solving the issue by scoring the conversation on a three-point scale.
A score of 1 denotes that the conversation was not helpful to the \texttt{asker} at all, and no progress was made; a score of 2 denotes some progress towards solving or diagnosing the issue, and a score of 3 indicates that the issue was solved. 
In cases where experienced \texttt{helpers} propose alternate solutions, success is measured by progress towards this \textit{new} goal.
Table~\ref{tab:success_aggregations} shows the overall statistics, and instructions for friction and success annotation can be found in the Appendix~\ref{sec:annotator-instructions}.
We obtain an agreement of $\alpha=0.58$ on success annotation as measured by Krippendorff's Alpha~\cite{castro-2017-fast-krippendorff}.

\begin{table}[t!]
\centering
\resizebox{\columnwidth}{!}{
\begin{tabular}{clp{8cm}l}
\toprule
\textbf{Turn} & \textbf{Speaker} & \textbf{Utterance} & \textbf{Grounding Act} \\
\midrule
0 & A (\texttt{asker}) & i have recently installed nvidia driver (working), but upon restart i get an error message: "failed to initialize nvidia kernel module" - anyone have any tips? :) & \\
1 & B (\texttt{helper}) & manf. drivers? & \\
\rowcolor{MutedOrange} 2 & A (\texttt{asker}) & sorry im not familiar with manf. drivers. i installed NVIDIA-Linux-x86-195.36.24-pkg1.run :) & \rr \\
\rowcolor{MutedOrange} 3 & B (\texttt{helper}) & yes i meant from nvidia site :) & \rep \\
\bottomrule
\end{tabular}
}
\caption{A typical conversation in our dataset, containing instances of \rr~and \rep~acts.}
\label{tab:nvidia-conversation}
\end{table}

\subsection{Measuring Friction Agreement}
\label{sec:agreement}
Measuring inter-rater agreement on friction detection is not straightforward, since we must account for agreement both in identifying an instance of friction \textit{and} the turn interval in which it occurs. 
To simplify this measurement, we compute overlap metrics for each pair of annotators, as in \citet{markowska-etal-2023-finding}. 
Agreement between an annotator pair is reported as the average of a modified version F1 score to measure interval overlap.
Specifically, for two annotators $A_1$ and $A_2$, we \textbf{average} two F1 scores---one treating annotations from $A_1$ as ground truth and those from $A_2$ as predictions and vice versa.\footnote{Our operationalization of F1 makes it asymmetric, hence $F1(A_1, A_2)$ is not guaranteed to be equivalent to $F1(A_2, A_1)$.}
We compute agreement in two different settings:

\paragraph{Friction Found.} In this \textit{relaxed} setting, an interval is ``found'' if \textit{any} turn within that friction window is part of \textit{any} predicted interval.
This setting does not require one-to-one mapping between predicted and gold friction instances.
Here, predicting one dialog turn within a gold friction interval is equivalent to predicting all turns correctly.

\paragraph{Friction Overlap.} We consider a second setting that rewards the \textit{degree} of overlap with the gold interval.
We first match each instance of friction with the predicted instance with the highest overlap, ensuring a one-to-one mapping between a predicted and gold friction interval.
For each \textit{matched} interval, we compute the Jaccard similarity between the two intervals, resulting in higher scores for predictions that better align with human-annotated instances of friction and penalizing predicting multiple short or overtly long instances.\footnote{This is similar in spirit to methods discussed in \citet{ortmann-2022-fine}, adapted for our task.}
A perfect score indicates an exact overlap between predicted and gold intervals.
We use these same two settings to compute model performance (Table \ref{tab:results}).
Table~\ref{tab:agreement_friction} shows agreement between pairs of annotators.
Given their high agreement, A1 and A2 annotated 80\% of the final dataset, while A3 annotated 20\%.

\input{tables/annotator-agreement}

\subsection{Annotating for Grounding Acts}\label{sec:grounding-acts}
Our annotations reveal that successful conversations contain less friction (Table \ref{tab:success_aggregations}). 
However, when friction is present, can participants collaboratively rebuild CG to complete tasks successfully? 
To better understand and model loss and repair of CG, we identify particular \textit{speech acts} associated with repair in friction intervals, as in \citet{levelt1983monitoring,heeman1994detecting,bohus2008sorry,bonialmaking}. 
For each friction interval $I$ in a conversation, we identify specific turns within $I$ expressing two grounding acts: \texttt{RequestRepair} and \texttt{Repair}~\cite{Traum1992}.
\rr~indicates whether a participant, spotting friction, \textit{explicitly} requests conversational repair from their partner.  
\rep~indicates whether friction was \textit{addressed} by either participant with a clarification (Table \ref{tab:nvidia-conversation}). 
 
Identifying these acts not only helps us determine whether participants recovered from friction, but also helps us to study in greater detail whether models can detect friction.
For example, this framework allows us to measure whether models detect friction only when in the presence of explicit requests or if they can identify \textit{implicit} cases of common ground misalignment.
This is important, as using LLMs as conversational partners or as mediators in human-human conversations depends on their ability to detect \textit{implicit} cases of friction. 

We sample 70 conversations containing 152 instances of friction to study the effects of grounding on task success. 
21 conversations received a success score of 1 (No Progress), 26 received a score of 2 (Some Progress), and 23 received a score of 3 (Success). 
Since conversations with friction tend to be longer, this sample has a higher average length than our dataset overall. 
Two authors annotated each friction instance in this subset for the presence or absence of \rr~and \rep acts, obtaining inter-rater scores of $0.69$ on \rr~, and $0.63$ on \rep~, measured using Cohen's Kappa~\cite{cohenkappa}.

%% file: tables/annotator-agreement.tex
\begin{table}[!ht]
\centering
\small
\begin{tabular}{c|ccc} 
\midrule
 & \multicolumn{1}{c}{\textbf{$A_1$}} & \multicolumn{1}{c}{\textbf{$A_2$}} & \multicolumn{1}{c}{\textbf{$A_3$}} \\ 
\midrule
\textbf{$A_1$} & -- & 65.91 / 25.86 & 48.0 / 18.21 \\
\textbf{$A_2$ }& -- & -- & 43.88 / 13.58 \\
\textbf{$A_3$} & -- & -- & -- \\
\bottomrule
\end{tabular}

\caption{Inter-rater agreement of detecting conversational frictions in Ubuntu-CG. Each cell contains the average of F1 scores between two annotators in two settings described in \S~\ref{sec:agreement} (Friction Found/Span Overlap).}
\label{tab:agreement_friction}
\end{table}

%% file: sections/40-human.tex
We study the relationship between the presence of conversational friction in goal-driven conversation and its success in Ubuntu-CG, and present our principal findings from the data below.

\paragraph{Successful conversations contain less friction.} In Ubuntu-CG, 61\% percent of conversations contained friction.
In contrast, of conversations where the \texttt{helper} succeeded in solving the \texttt{asker}'s issue (receiving a score of 3), only 54.5\% contained friction (Table \ref{tab:success_aggregations}). 
Conversations where participants make some progress or succeed contain less friction on average as compared to conversations where they did not make \textit{any} progress, as the former exhibits some amount of grounding effort by the participants (Table~\ref{tab:success_aggregations}, Column 4).
This is further supported by the proportion of unaddressed repair efforts (Table \ref{tab:conversations_repairs}, Column 4).

\paragraph{Friction is more likely in longer conversations.} While conversation length shows no clear pattern with task success (Table \ref{tab:success_aggregations}), the mean length of a conversation containing friction is 49 (median 55), as compared to an average of 29 (median 22) of those without friction. Compared to the overall mean length of the dataset 40.56 (median 33), it is plausible that conversational friction and repair through the process of grounding contribute to the increased number of turns it takes to complete the conversation.

\subsection{Role of Grounding Acts in Task Success} 

Conversations where participants could not make \emph{any} progress towards diagnosing a particular issue (success score of 1) are characteristically different from conversations receiving a score of 2 or 3. 
In a retrospective study, we analyze the presence of grounding acts (\rr~and \rep) in conversations that received a score of 1 (No Progress) as compared to conversations receiving a score of 2 or 3 (Some Progress or Success). 

\input{tables/progress-ground}

We focus on the proportion of \rr~acts that were not addressed.
This captures instances of friction where, despite one participant spotting a potential mismatch in common ground, their efforts are not reciprocated by their conversational partner. 
Notably, conversations with no progress exhibited a higher proportion of these unacknowledged \rr~acts  (Column 4 in Table \ref{tab:conversations_repairs}). 
This further shows that achieving a communicative goal requires \textit{both} participants to engage in grounding.

%% file: tables/progress-ground.tex
\begin{table}[t!]
\scriptsize %
\setlength{\tabcolsep}{4pt} %
\centering
\resizebox{\columnwidth}{!}{%
\begin{tabular}{lccc}
\toprule
\textbf{\shortstack{Degree of \\ Progress}} & \textbf{\#Convs} & \textbf{\shortstack{Instances \\ (Repair/ReqRepair)}} & \textbf{\shortstack{Unaddressed \\ ReqRepair (\%)}} \\ 
\midrule
2 or 3 & 49 & 102 (83/75) & 22.67\\ 
\midrule
No Progress (1) & 21 & 50 (38/36) & 30.56\\ 
\bottomrule
\end{tabular}%
}
\caption{Summary of success and grounding acts in our analysis subset of 70 conversations. In conversations with no progress, more requests for repairs go anaddressed.}
\label{tab:conversations_repairs}
\end{table}

%% file: sections/50-llm.tex
Identifying friction in ongoing conversations is a first step towards analyzing the content of the CG. 
After establishing simple finetuned baselines on the task of conversational friction detection, we go on to explore whether larger LLMs can identify and explain instances of conversational friction in Ubuntu-CG.

\input{tables/results}

\subsection{Experimental Setup}\label{sec:eval-metrics}

\paragraph{Encoder-Only Baseline.} Before moving on to prompting, we first explore a baseline setting in which we finetune a small encoder-based model \texttt{distilroberta-base}~\cite{Sanh2019DistilBERTAD} on five randomly-split folds of Ubuntu-CG (Appendix \ref{appendix:finetuning}).
Given an excerpt of a conversation consisting of a target turn $t$ and a context window of $k$ turns before and after $t$, the model is trained to predict whether $t$ is part of an annotated instance of friction.
We report results on context windows of three and five, and notice no significant improvement in using a higher $k$. 
Finetuning a lighter-weight model enables us to understand the extent to which friction can be identified from surface features.  

\paragraph{Decoder-Only Models.} Given a \textit{full} conversation, we prompt several larger decoder-only LLMs to output \textit{a list of turn intervals} exhibiting friction (Prompt~\ref{prompt:friction-intervals} in Appendix).\footnote{We experimented with several prompting strategies such as adding random exemplars, self-consistency, and chain-of-thought reasoning, but found that they did not beat the F1 scores obtained simply by asking the model to detect friction windows along with brief explanations of why a dialog window represents friction.}
For each predicted turn interval, the model must provide a brief explanation for the cause of friction. 
We also include a setting where we provide models with the elaborations of technical terminology as outlined in \S\ref{sec:friction-annotation} (``w\ Elab''). Note that in this setting, an LLM predicts possible friction intervals in a single pass, in contrast to the encoder baseline, where models make predictions on every single turn separately, taking neighboring turns as context.

\paragraph{Evaluation Metrics.}
We evaluate models in the \textbf{Friction Found} and \textbf{Friction Overlap} settings~\ref{sec:agreement}.
While \textbf{Friction Found} allows models like \llamasmall~\cite{touvron2023llamaopenefficientfoundation} to obtain high recall scores by over-predicting friction intervals, \textbf{Friction Overlap} penalizes this behavior.
For all experimental settings, we set temperature to $0.01$.

\subsection{Results}

The F1 scores of our baseline encoder-only models (Table~\ref{tab:results}) in two settings (context window $k=3$ and $k=5$) give us an estimate of the degree to which friction is identifiable from shallower, local features as opposed to more complex, contextual, and implicit cases of friction like subtle clarification questions or other pragmatic phenomena.
Error analysis reveals that instances predicted correctly by our baselines often contained explicit markers of friction, such as a user expressing frustration or dissatisfaction. 
While these baselines were competitive with some decoder-only models, especially under macro-averaging (Table \ref{tab:macro-results} in Appendix) which weights individual friction instances equally regardless of conversation, the lack of model explanations limits deeper interpretation beyond error analysis.
However, it does indicate that our dataset contains learnable surface features such as explicit expressions of frustration or anger that can be captured by a small encoder-only model.

Under both evaluation settings, \gpt \ without any further technical elaborations obtained the highest F1 score. 
This setting is used for all further error analysis and ablations.
Models over-predict friction intervals (see \#Predictions in Table \ref{tab:results}).

\paragraph{The effect of \texttt{gpt-4o} Elaborations.} Explaining technical terms with \texttt{gpt-4o} helped our human annotators better understand the flow of information in a conversation.
However, in the relaxed evaluation setting of \textbf{(Friction Found)}, adding elaborations does not improve prediction scores of models. 
For \llamasmall~and \texttt{gpt-4o-mini}, adding elaborations improves recall and hence the overall F1 score. 
This may be due to elaborations ``sharpening'' the predicted intervals.

\begin{table}[h!]
\scriptsize %
\setlength{\tabcolsep}{4pt} %
\centering
\resizebox{\columnwidth}{!}{%
\begin{tabular}{lcc}
\toprule
\textbf{Model} & \textbf{\shortstack{Success Prediction \\ (Spearman's $\rho$)}} & \textbf{\shortstack{Binary Friction \\ Presence (Cohen's $\kappa$)}} \\
\midrule
\gpt & \textbf{0.776} & \textbf{0.380} \\
\rowcolor{MutedBlue} \gpt~w/ Elab. & 0.743 & 0.310 \\
\gptmini & 0.699 & 0.205 \\
\rowcolor{MutedBlue} \gptmini~w/ Elab. & 0.634 & 0.205 \\
\llamasmall & 0.261 & 0.193 \\
\rowcolor{MutedBlue} \llamasmall~w/ Elab. & 0.235 & -0.249 \\
\llamabig & 0.702 & 0.290 \\
\rowcolor{MutedBlue} \llamabig~w/ Elab. & 0.630 & 0.223 \\
\bottomrule
\end{tabular}%
}
\caption{Spearman's $\rho$ and Cohen's $\kappa$ for the related tasks of predicting success friction presence. Models align more with humans on the success of a conversation.}
\label{tab:ablation}
\end{table}

\paragraph{Ablations.}
Human annotators do not always agree on the location of 
friction and repair-related grounding acts. 
To understand whether models can make binary judgments as to whether or not friction is present without identifying their location, we prompt models to predict the presence of friction \textit{without} pinpointing specific dialog turns. 
This allows us to assess the model's ability to predict friction as a broader phenomenon. 
We also evaluate the capability of models to predict the \textit{success} of the task undertaken in the conversation on a three-point scale, as in \S\ref{sec:task-success}. 

We evaluate the binary prediction task with Cohen's $\kappa$, framing it as inter-rater agreement between models and humans. 
Models' over-prediction of friction intervals persists in the conversation level as well (Table~\ref{tab:ablation}).
Predictions on task success, on the other hand, is highly correlated with annotator ratings of success.

%% file: tables/results.tex
\begin{table*}[h]
    \centering
    \small
    \begin{tabular}{lccccccc}
        \toprule
        & \multicolumn{3}{c}{\textbf{Friction Found}} & \multicolumn{3}{c}{\textbf{Friction Overlap}} & \multirow{2}{*}{\textbf{\#Predictions}} \\
        \cmidrule(lr){2-4} \cmidrule(lr){5-7}
        \textbf{Model} & \textbf{Precision} & \textbf{Recall} & \textbf{F1} & \textbf{Precision} & \textbf{Recall} & \textbf{F1} & \\
        \midrule
        \gpt & 31.50 & 43.69 & \textbf{34.01} & 13.50 & 18.74 & \textbf{14.61} & 495 \\
        \rowcolor{MutedBlue} \gpt~w/ Elab. & 31.63 & 37.46 & 32.22 & 13.54 & 16.59 & 14.00 & 435 \\
        \gptmini & \textbf{32.75} & 27.86 & 28.01 & \textbf{13.67} & 12.32 & 12.10 & 316 \\
        \rowcolor{MutedBlue} \gptmini~w/ Elab. & 28.54 & 28.67 & 26.51 & 13.63 & 14.11 & 12.81 & 392 \\
        \llamasmall & 16.72 & 47.28 & 22.53 & 6.87 & 18.72 & 9.14 & \textbf{1282} \\
        \rowcolor{MutedBlue} \llamasmall~w/ Elab. & 15.98 & 46.33 & 21.73 & 7.11 & \textbf{20.02} & 9.58 & 1253 \\
        \llamabig & 21.70 & \textbf{48.09} & 27.97 & 8.93 & 20.26 & 11.59 & 857 \\
        \rowcolor{MutedBlue} \llamabig~w/ Elab. & 16.72 & 39.83 & 22.06 & 7.35 & 16.76 & 9.52 & 959 \\
        \texttt{distilroberta-base}~($k=3$, finetuned) & 14.52 & 26.32 & 16.37 & 5.93 & 10.12 & 6.57 & - \\
        \texttt{distilroberta-base}~($k=5$, finetuned) & 13.03 & 26.14 & 15.52 & 6.38 & 11.84 & 7.43 & - \\
        \bottomrule
    \end{tabular}
    \caption{Micro-averaged precision, recall, and F1 scores of different models on detecting friction. Each conversation contributes equally to the final metrics regardless of the number of friction instances it contains. \#Predictions refer to the total number of instances of conversational friction found by each model. For reference, annotators identified \textbf{238} instances in total. \gpt~\emph{without} Elaboration of technical terms (Sec \ref{sec:friction-annotation}) performed best across all models. Macro-averaged results can be found in Table~\ref{tab:macro-results} in the Appendix.}
    \label{tab:results}
\end{table*}

%% file: sections/55-error_analysis.tex
We now investigate the successes and failures of \texttt{gpt-4o}, the strongest performing model at this task. 

\paragraph{Undetected frictions are deeper in conversations.}
As a conversation proceeds, detecting friction requires a deeper understanding of preceding turns. 
To explore whether the \textit{position} of friction impacts model accuracy, we stratify our results by conversational depth and calculate the relative depth of each instance of friction as the ratio of the first turn of the friction interval to the conversation length multiplied by 100. 
The mean relative depth of a detected instance of friction ($35.19$) is significantly smaller than the mean relative depth of a detected instance ($49.62$), according to an independent t-test ($p<0.01$). 
This indicates that models struggle with taking a longer context into account while determining whether participants' versions of common ground are misaligned. 

\paragraph{Implicit cases of friction are harder to detect.} 
Models, particularly \texttt{gpt-4o}, are more likely to correctly identify friction when an explicit request for conversational repair is present.
Specifically, 77.22\% of detected frictions involved an explicit \rr, compared to $64.81$\% of frictions that went undetected ($p < 0.05$). 
This highlights the tendency of models to rely on overt cues that signal a common ground misalignment.

Consider the conversation in Table \ref{tab:nmap-conversation}. 
A's utterance \textit{``how about nmap''} (Turn 22) is not introducing \texttt{nmap} as an option, but following up on B's earlier suggestion in Turn 21 by asking \textit{how} \texttt{nmap} can be used to solve the issue. 
B reveals that they did not understand this interpretation through their response in Turn 23 (\textit{``yeah, i said nmap,''}), prompting A to issue a \rep~act.
We hypothesize that this unconventional way of issuing a \rep~(through a question) without an explicit \rr~results in an undetected conversational friction.

\begin{table}[t!]
\centering
\resizebox{\columnwidth}{!}{
\begin{tabular}{clp{8cm}}
\toprule
\textbf{Turn} & \textbf{Speaker} & \textbf{Utterance} \\
\midrule
16 & B (\texttt{helper}) & btw, you do need to restart the ssh server for it to work on the new ip(s) \\
17 & A (\texttt{asker}) & sudo service ssh restart? \\
18 & B (\texttt{helper}) & yeah \\
19 & A (\texttt{asker}) & is the service ssh or anything else? \\
20 & B (\texttt{helper}) & yep thats the service \\
21 & B (\texttt{helper}) & and you can check if its listening with nmap \\
\rowcolor{MutedOrange} 22 & A (\texttt{asker}) & how about nmap? \\
\rowcolor{MutedOrange} 23 & B (\texttt{helper}) & yeah, i said nmap \\
\rowcolor{MutedOrange} 24 & A (\texttt{asker}) & I mean how do I use nmap to find that out? \\
\bottomrule
\end{tabular}
}
\caption{A conversation showing an undetected case of friction, where a \rep~act is expressed through a question (Turn 24). B misinterprets A's question in Turn 22 as a suggestion, while, as revealed in Turn 24, A was simply following up on B's early suggestion of using \texttt{nmap} from Turn 21.}
\label{tab:nmap-conversation}
\end{table}

\paragraph{Comparing Model and Human Explanations.} Collecting model explanations along with friction interval predictions allows us to evaluate whether they accurately capture the cause of friction. 
In most cases, this amounts to correctly pointing out the cause of misalignment in participants' respective versions of the CG.
To study this, we ask two non-author Linux experts to annotate the similarity between gold friction explanations and model-generated explanations for 64 instances of friction on a three-point scale, marking them as either (1) dissimilar, (2) somewhat similar, or (3) equivalent.
Agreement between the two annotators using Spearman correlation is $\rho=0.61$, with $p<0.01$.

Most model explanations accurately captured the cause of friction, with $57.81$\% instances annotated as equivalent and $34.37$\% as somewhat similar.
Only $7.8$\% explanations did not point out the cause of friction at all. 
Echoing our earlier findings, models struggle to pinpoint the cause of friction \textit{even when they identified the window correctly}.
For example, in Figure \ref{fig:gpt4-fail}, Row 1, A mistakenly assumes that B's suggestion for a command (Turn 25) includes the keyword ``try'', leading to the error ``try command not found'' (Turn 32) after which B repeats the \texttt{dmesg} command removing ``try'' at the beginning. 
LLMs must be able to pinpoint causes of friction to issue repairs that address the friction directly, a crucial ability in settings where LLMs are used for dispute resolution~\cite{tan2024robotsmiddleevaluatingllms}.

%% file: sections/65-related_work.tex
\section{Related Work}\label{sec:related-work}
The speech-act based approach in \citet{Traum1992} has been used to study cooperative grounding acts in the Meetup~\cite{ilinykh2019meetup} and Spot the Difference~\cite{lopes-etal-2018-spot} datasets~\cite{mohapatra-etal-2024-conversational}. 
While conversations in such scenarios also require grounding, both datasets involve conversational participants interacting in a \textit{physical} setting.
Because of the additional modality, the mutually shared basis of their CG (e.g. an object both or one participant can see) is not available to the reader, making it difficult to capture what causes friction from text alone.

\citet{markowska-etal-2023-finding} track speaker versions of the CG through speaker ``beliefs'' expressed in conversations in the LDC Callhome \cite{callhome1997} corpus. 
However, participants are not as incentivized to build and maintain a rich CG since the conversations are not goal-driven, and are between close friends or family.
\citet{khebour-etal-2024-common} annotate a task-oriented corpus for multi-modal features and dialogue moves to model shared beliefs and questions under discussion. 
The authors train LSTM-based classifiers of dialogue moves relevant to tracking CG, finding that utterances may or may not be aligned with other modalities such as gesture. 
This highlights the challenge of tracking common ground in physically situated dialogue; our dataset simplifies the focus to text alone. 

\citet{shaikh-etal-2024-grounding} use grounding acts to compare the degree of grounding by LLMs in human-LLM conversations, finding that LLMs perform less computational grounding. 
Our work complements these directions by focusing on whether LLMs can \textit{detect} when and how participants in a conversation might lose track of CG. 
In more recent work, ~\citet{shaikh2025navigatingriftshumanllmgrounding} focuses on this divergence across several grounding acts, and introduces an annotated dataset of LLMs failing to ground in human-LLM conversations.

Also related to our work is the concept of positive friction as explored in ~\citet{i̇nan2025betterslowsorryintroducing}. 
Here, the authors view ``friction'' not as a misalignment in common ground that detracts from the goal of the conversation, but as a series of communicative ``movements'' such as pausing, revealing speaker assumptions, etc., that facilitate long-term success over short-term progress in a conversation.

%% file: sections/70-conclusion.tex
\section{Conclusion and Future Work}
In this case study, we have conducted what is to our knowledge the first investigation of friction and repair of CG for task-oriented dialogue in a real-world, text-only setting.
Our qualitative and quantitative results reveal that friction in goal-oriented dialog is inevitable, and it takes effort from \textit{both} participants to repair the CG to make progress towards a task. 
Keeping track of CG over text is no easy feat---it requires participants to be vigilant about implicit cues in text that might signal a potential misalignment.  
While some \texttt{helper}s in our dataset anticipated and prevented potential friction or issued \rep~acts once friction did happen, LLMs such as \gpt~struggled with detecting and explaining cases of friction in the absence of explicit evidence. 

As LLMs are deployed in settings such as education~\cite{wang2024largelanguagemodelseducation}, future work might explore improving their ability to understand \textit{implicit} ruptures in CG---an LLM tasked with analyzing conversations between a student and teacher should be able to detect the loss of common ground for better learning outcomes.
Another future direction involves explicitly \emph{modeling} the CG, consisting of propositions that are part of participants' underlying mental state. 
Recent work has demonstrated that LLMs draw plausible inferences about such propositions in non-conversational settings \citep{hoyle-etal-2023-natural}. 
Conceptually, thought bubbles such as those illustrated in Figure~\ref{fig:teaser} could be populated automatically, operationalizing the detection of CG misalignments by similarity-based comparison and contrast of participants' individual belief spaces.

%% file: sections/75-limitations.tex
Our study takes an important step towards quantifying the role of grounding in goal-oriented dialog and studying LLM capabilities of detecting friction.
Unlike studies that simulate conversations between participants in artificial settings to gain access to their mental states and the common ground, we do not have access to conversational participants' common ground or mental states beyond what is expressed in the text conversation.
In addition, we do not have access to the degree of self-effort that goes into solving an issue alongside a conversation---the \texttt{asker} might simultaneously have been searching the internet for answers while engaged in conversation.

Another limitation of our work stems from the fact that we limit our analyses of LLMs to their role as an observer of human-human conversation, and not as a participant. 
Given that LLMs perform differently in linguistic tasks (such as responding to a query) as opposed to metalinguistic tasks (spotting a mismatch in common ground), it is possible that a model \emph{response} addresses a mismatch in common ground indirectly as a conversational partner while failing to identify the cause of friction as an observer~\cite{hu2024auxiliarytaskdemandsmask}.
However, we believe that understanding LLM behavior of tracking common ground is an essential prerequisite to many other downstream research questions, such as cases where an LLM is used as a conversational facilitator~\cite{Argyle2023}.

Although the conversations take place purely through text, participants sometimes shared links to blog posts and tutorials, many of which now no longer work. 
In rare cases, it might be possible that the cause (or resolution) of a friction instance is rooted in such a link. 
We also do not have access to their screens or other metadata about the user that might have been instrumental in resolving friction.

\section*{Acknowledgments}
We thank our anonymous reviewers for their comments and suggestions throughout the reviewing process. We also thank Atrey Desai, Nyle Masood, Ayush Kumar, Maya Srikanth, and Pratheek Hegde for providing valuable annotations. This work was supported in part by the US National Science Foundation award 2124270. Any opinions, findings, conclusions, or recommendations expressed in this material are those of the author and do not necessarily reflect the views of the National Science Foundation.

%% file: sections/80-appendix.tex
\section{Appendix}

\subsection{Further Annotation Details}
In our dataset of 200 conversations, we divide annotations into 10 batches, each containing 20 conversations.
Three of those batches (60 conversations) were annotated by all annotators, going towards computing the inter-rater agreement scores seen in Table~\ref{tab:agreement_friction}. 
The rest of the conversations were annotated by a single annotator.
For the conversations annotated by all three annotators, we picked annotations from either A1 or A2, since they had stronger inter-rater agreement. 
Overall, out of the 10 batches, 4 batches were annotated by A1, 4 were annotated by A2, and 2 were annotated by A3.

The model scores and the inter-rater agreement values are calculated using the same metrics, friction-found and friction-overlap~\ref{sec:agreement}.
They are comparable with the caveat that results in Table 6 are unidirectional (models against gold data), and those in Table 4 are made two-sided (by averaging $A_i$ against $A_j$ and $A_j$ against $A_i$). 
We note that while annotator agreement numbers are modest, the best inter-rater agreement between our annotators (A1 and A2) is significantly higher than any of our considered models. These two particular annotators (A1 and A2) annotated 80\% of our dataset.

\subsection{Model Finetuning Details}
\label{appendix:finetuning}
For our finetuning experiments, we create five random train-test splits of our dataset, with 30\% of conversations in Ubuntu-CG in each split going towards the ``test'' set.
For each of our context windows ($k=3$ or $k=5$ turns), we train a \texttt{distilroberta-base} model with 82M parameters for 15 epochs with a learning rate of $4e-5$.
As evaluation data, we pick a single fold and isolate a part of its training set as our development data.
Since this is a class-imbalanced dataset (turns not containing friction greatly outnumber turns containing friction 10 to 1), we artificially reduce the number of negative samples in our \textit{training} data, however resampling was not done on the test data. 
We also experimented with larger encoder-only models, but they all performed much poorer than \texttt{distilroberta} under both full-parameter and classification head-only training. 

\paragraph{Relationship with NLI.} The setup we choose for our finetuned baselines may resemble that of classical textual entailment, where, given a premise sentence, a model must determine whether a hypothesis sentence is entailed, contradicted, or neutral in relation to the premise. 
Friction detection in this baseline setting \textit{could} be \textit{recast} as an NLI-style task~\cite{white-etal-2017-inference}.
We do not experiment with existing finetuned NLI models since, as is, our task in the baseline setting does not cleanly or directly map to the task of textual entailment in its current form, and hence, would produce unreliable labels.

\subsection{Computational Details}
The \llamabig~models were used with 4bit quantization to fit on two A6000 GPUs.
 
\subsection{Prompts}
We outline all prompts used in the paper below. In the interest of presentation, they are broken into modules. For example, Prompt~\ref{prompt:friction-intervals} and Prompt~\ref{prompt:io-format} would combine to form a single prompt for friction detection, and Prompt~\ref{prompt:explanations} is plugged in the middle to make use of \gpt-generated explanations.

\begin{prompt}[title={Prompt \thetcbcounter: Friction Detection Prompt}, label=prompt:friction-intervals]
\texttt{\colorbox{MutedBlue}{Prompt:} \#\#\# TASK DESCRIPTION: Detecting "Conversational Friction" in Online Conversations.\\ \\ 
Given a conversation between two participants in an online chat forum, label one or more turns in the conversation where there is evidence of friction between the two participants, that is, where they don’t seem to fully understand each other or seem to not be on the same page. This friction could be due to a mismatch between their goals, due to a false assumption one participant made about the other leading to a misunderstanding, and so on. These may result from a mismatch in the common ground between the two participants.
\\
\\
A strong indicator of conversational friction could be a participant asking the other participant to revisit or clarify previously shared content in the conversation, in a process known as conversational repair. However, in many cases there may not be an explicit Repair Request issued by a participant but from context it can be reasoned that a participant is struggling to keep up with the conversation. In some cases, it becomes apparent that a participant was requesting conversational repair in a turn only after reading through subsequent turns. In that case, go back and annotate that turn as friction.
\\
\\
Note that possible friction can occur in a single turn (in which case, mark that specific turn), or through a series of turns (in which case, mark the window of turns that all together add up to a repair request). In each of these cases, you should mark the turn(s) where the friction is most apparent. Also write a brief explanation of why you think that turn is an instance of conversational friction as defined above. 
}
\end{prompt}

\begin{prompt}[title={Prompt \thetcbcounter: Input/Output Format}, label=prompt:io-format]
\texttt{\colorbox{MutedBlue}{Prompt:} \#\#\# INPUT: \\
Conversation:
\{convo\_text\}
\\
\\
Now, follow the output format below to annotate the conversation.
\\
\\
\#\#\# OUTPUT FORMAT:
\\
\\
First output the turns showing conversational friction in a dictionary. If there is more than one instance of friction, list them in the order they appear in the conversation. If there's no friction in the conversation, set "friction\_present" to false and don't provide any other fields.
\\
Follow the output format below to annotate the conversation.
\\
\\
\{\{ \\
    "friction\_present": [Choose true or false], \# if false, stop here \\
    "friction1": [X, Y], \# the start and end turns of the first instance of friction \\
    "explanation1": "Brief explanation for friction1", \\ 
    "friction2": [X, Y], \# If there is more than one instance of friction \\
    "explanation2": "Brief explanation for friction2" \\ 
.... \\
\}\}
}
\end{prompt}

\begin{prompt}[title={Prompt \thetcbcounter: Adding Explanations}, label=prompt:explanations]
\texttt{\colorbox{MutedBlue}{Prompt:} \#\#\# EXPLANATIONS \\
\\ 
To clarify the many technical terms used in the conversations, you are also provided an explanation of terms used in a particular turn at the end of the turn. This explanation is provided in the format: Turn X Explanation: <Explanation of the terms used in Turn X>. In general, the format of the conversation is as follows:
\\ \\ 
**[Turn 0] User A:** <Message about current current issue with linux> \\
Turn 0 Explanation: <Contextual explanation of the technical terms used in the conversation> \\
**[Turn 1] User B:** <Response to Turn 0> \\ 
Turn 1 Explanation: <Contextual explanation of the technical terms used in the conversation> \\ 
... \\
... \\
\\ \\ 
**NOTE:** In addition to the conversation, optionally use the explanations provided to better understand what's going on in the conversation. Discard the explanations if you feel they are not necessary.  \\
}
\end{prompt}

\begin{prompt}[title={Prompt \thetcbcounter: Success Prediction}, label=prompt:success-prediction]
\texttt{\colorbox{MutedBlue}{Prompt:} \#\#\# TASK DESCRIPTION \\
You will be given a conversation between two participants A (usually the **user** seeking help) and B (usually the **helper**) who are trying to solve an issue in Ubuntu together on the \#Ubuntu IRC channel. Your task is to determine how successful the conversation was towards resolving the issue of the user. 
\\ \\ 
Mark how helpful the conversation was to whoever was asking for help on a scale of 1-3, where each number on the scale has the following meaning: 
\\ \\ 
- 1 (NO PROGRESS): This indicates that the conversation was not helpful to A at all in resolving their issue, and they did not make any progress towards solving the problem. \\
- 2 (SOME PROGRESS): This indicates that the participants made some progress towards solving the problem. They might not have resolved the issue entirely, but they made progress in diagnosing the problem or solved a part of the problem. \\
- 3 (SUCCESS): This indicates that the participants solved the problem they initially set out to solve, or the problem that evolved in the course of the conversation. \\ 
\\ 
The scores hold true even if they themselves realize the issue in the course of the conversation and proceed to solve it. It also holds true even if the conversation went off-topic, as long as the participants were able to solve the problem at hand. \\ 
\\ 
NOTE: The problem that A starts the conversation with might not be the right problem to solve at all, and the helper (usually B) might suggest what the right issue to solve is. In that case, solving the re-defined problem will decide conversational success on this scale.  \\ 
\\ 
\#\#\# INPUT \\
\\ 
Conversation: \\
\\ 
\{convo\_text\} \\ 
\\  
\#\#\# OUTPUT \\
\\
First, provide the success score for the conversation on a scale of 1-3. Then, provide a brief explanation explaining the score in the format below: \\
\{\{ \\
    "success\_score": [1/2/3] \# 1 for NO PROGRESS, 2 for SOME PROGRESS, 3 for SUCCESS. Output score only \\
    "explanation": "Brief explanation for the success score" \\ 
\}\} \\
}
\end{prompt}

\begin{prompt}[title={Prompt \thetcbcounter: Binary Friction Detection}, label=prompt:friction-binary]
\texttt{\colorbox{MutedBlue}{Prompt:} \#\#\# TASK DESCRIPTION: Detecting "Conversational Friction" in Online Conversations\\ \\ 
Given a conversation between two participants in an online chat forum, output whether there is evidence of conversational friction between the two participants. Conversational friction occurs when participants in a conversation don’t seem to fully understand each other or seem to not be on the same page. This friction could be due to a mismatch between their goals, due to a false assumption one participant made about the other, leading to a misunderstanding, and so on. These may result from a mismatch in the common ground between the two participants.
\\ \\ 
A strong indicator of conversational friction could be a participant asking the other participant to revisit or clarify previously shared content in the conversation, in a process known as conversational repair. However, in many cases, there may not be an explicit Repair Request issued by a participant, but from the context, it can be reasoned that a participant is struggling to keep up with the conversation. 
\\ \\ 
NOTE: Friction is often signaled by the helpee asking a follow-up question. However, not all follow-up questions indicate that the speakers are not on the same page. For example, clarification questions that ask for information not assumed by either user to be in the common ground are not cases of conversational friction. Clarification questions that move the conversation forward without questioning the common ground are not cases of conversational friction. If there is **no conversational friction** make sure to indicate that in the output by setting "friction\_present" to false. 
\\ \\ 
\#\#\#  TASK: \\
\\
Given a conversation, list whether conversational friction occurs or not.
}
\end{prompt}

\subsection{Elaborations}
Examples of elaborations can be found in Table~\ref{tab:elaboration}.

\begin{table*}[h]
    \centering
    \footnotesize
    \begin{tabular}{p{0.40\textwidth}  p{0.45\textwidth}  c}
    \toprule
    \multicolumn{1}{c}{\textbf{Utterance}} & \multicolumn{1}{c}{\textbf{GPT Elaboration}} & \multicolumn{1}{c}{\textbf{Year}} \\ 
      \hline
       \small hi, i have ubuntu dapper and want to do a clean install of feisty using the live cd (I want to put feisty in my current ext3 partition and format ext3). When the installation process comes to the part about partitioning, (Erase hard disk, automatic, or manual), should I choose manual and if so, will there be a way to format ext3 and will it allow me to put feisty in my current ext3 partition without making a new & \small \texttt{Ubuntu Dapper and Feisty are code names for older versions of the Ubuntu operating system, specifically 6.06 (Dapper Drake) and 7.04 (Feisty Fawn), respectively. A 'live CD' allows you to run Ubuntu directly from the CD without installing it on your hard drive. 'ext3' is a type of file system used in Linux for organizing and storing files on a partition.} & \small 2005 \\
       \hline
        \small does passwords and encryption keys support hkps? & \small \texttt{"HKPS" stands for HTTP Keyserver Protocol Secure. It is a secure version of the HTTP Keyserver Protocol (HKP) used to retrieve encryption keys from a keyserver over a secure, encrypted connection. In the context of Ubuntu or other operating systems, this might refer to the secure retrieval or management of encryption keys, potentially in relation to applications or services that require encryption.} & \small 2010 \\ 
        \bottomrule

    \end{tabular}
    \caption{Explanation of technical terms present in dialog turns explained by GPT4. These help our annotators understand terms such as ``khps'', ``dapper'', or ``feisty''.}
    \label{tab:elaboration}
\end{table*}

\subsection{Annotator Instructions}\label{sec:annotator-instructions}

\begin{figure}[ht!]
    \centering
    \includegraphics[width=0.5\textwidth]{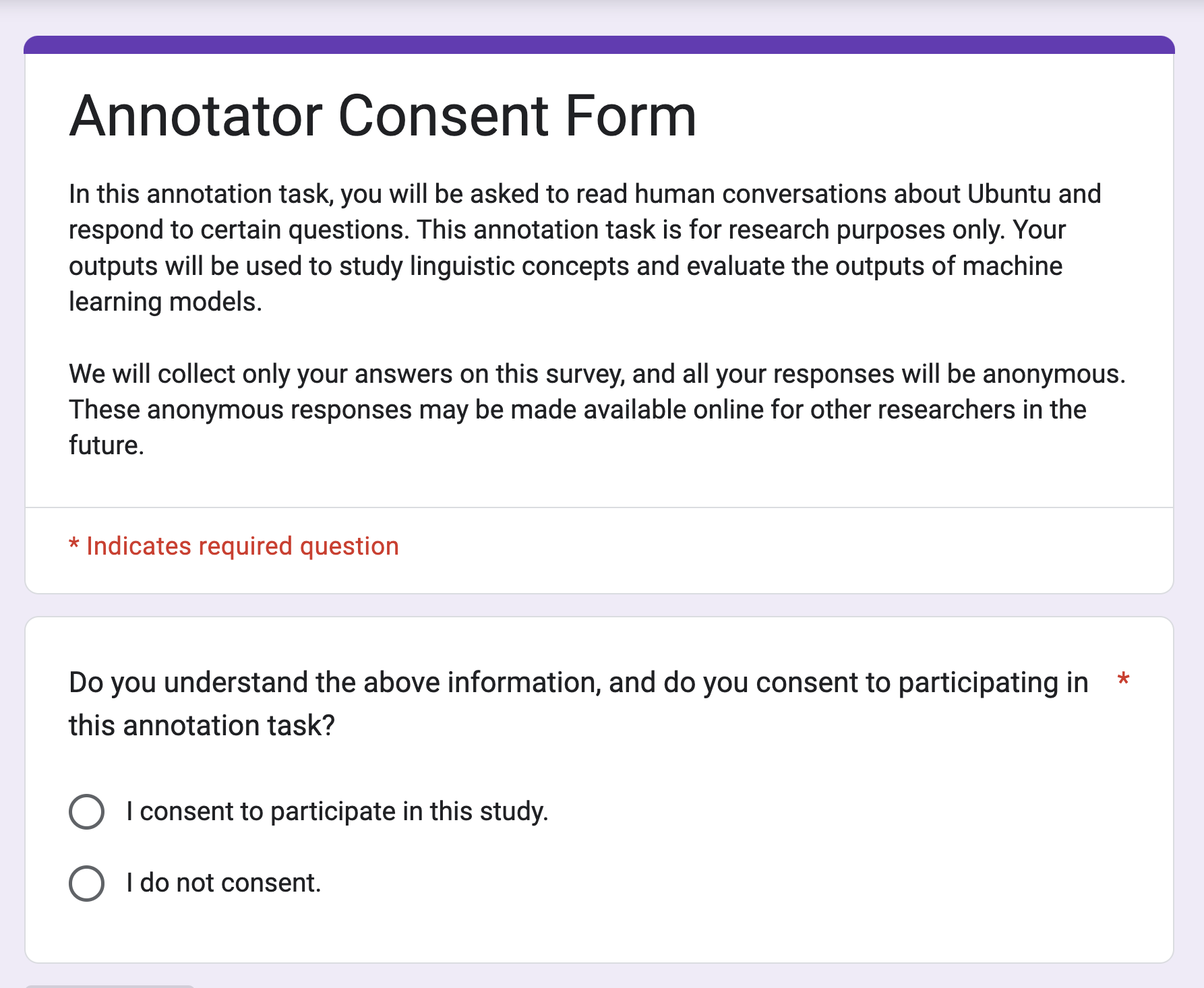}
    \caption{The consent form shown to annotators before each task.} 
  \label{fig:consent}
\end{figure}

\begin{figure*}[htb]
    \includegraphics[width=0.9\textwidth]{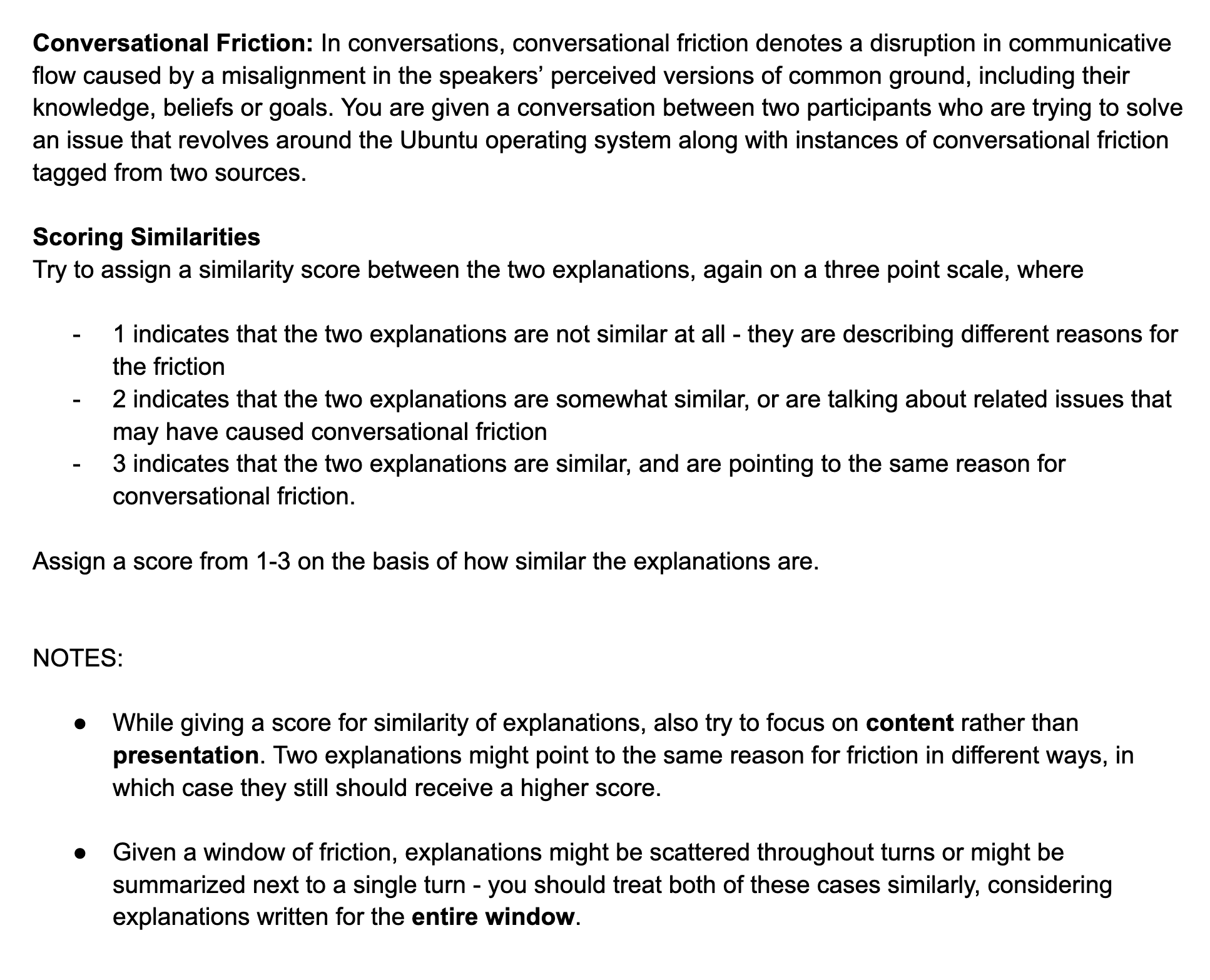}
    \caption{Instructions provided to the annotators for judging the similarity of \gpt and human-generated explanations for frictions. The annotators did not know the source of an explanation.} 
  \label{fig:similarity-instruction}
\end{figure*}

Before any annotation task, annotators had to fill out a consent form (Figure \ref{fig:consent}). 
To ensure we're measuring equivalent constructs, the annotator instructions was kept identical to Prompt~\ref{prompt:friction-intervals}. A more detailed instruction document can be found in the supplementary material.
The similarity scoring prompt is shown in Figure~\ref{fig:similarity-instruction}. 

\subsection{Breakdown of Table~\ref{tab:conversations_repairs}}
\begin{table}[t!]
\scriptsize
\setlength{\tabcolsep}{4pt}
\centering
\resizebox{\columnwidth}{!}{%
\begin{tabular}{lccc}
\toprule
\textbf{\shortstack{Degree of \\ Progress}} & \textbf{\#Convs} & \textbf{\shortstack{Instances \\ (Repair/ReqRepair)}} & \textbf{\shortstack{Unaddressed \\ ReqRepair (\%)}} \\ 
\midrule
3 & 23 & 45 (35/33) & 27.27 \\ 
2 & 26 & 57 (48/42) & 19.05 \\ 
\midrule
No Progress (1) & 21 & 50 (38/36) & 30.56 \\ 
\bottomrule
\end{tabular}%
}
\caption{Full breakdown of summary of success and grounding acts in our analysis subset of 70 conversations. In conversations with no progress, more requests for repairs go unaddressed.}
\label{tab:conversations_repairs_all}
\end{table}

\subsection{Macro-Averaged Results}\label{sec:macro-appendix}

For completeness, we also present the macro-averaged result in Table \ref{tab:macro-results}. 

\input{tables/results_macro}

%% file: tables/results_macro.tex
\begin{table*}[h]
    \centering
    \small
    \begin{tabular}{lccccccc}
        \toprule
        & \multicolumn{3}{c}{\textbf{Friction Found}} & \multicolumn{3}{c}{\textbf{Friction Overlap}} & \multirow{2}{*}{\textbf{\#Predictions}} \\
        \cmidrule(lr){2-4} \cmidrule(lr){5-7}
        \textbf{Model} & \textbf{Precision} & \textbf{Recall} & \textbf{F1} & \textbf{Precision} & \textbf{Recall} & \textbf{F1} & \\
        \midrule
        \gpt & 34.34 & 71.43 & \textbf{46.38} & 14.28 & 29.70 & \textbf{19.29} & 495 \\
        \rowcolor{MutedBlue} \gpt~w/ Elab. & \textbf{34.94} & 63.87 & 45.17 & \textbf{14.75} & 26.96 & 19.07 & 435 \\
        \gptmini & 32.28 & 42.86 & 36.82 & 13.97 & 18.55 & 15.94 & 316 \\
        \rowcolor{MutedBlue} \gptmini~w/ Elab. & 28.32 & 46.64 & 35.24 & 14.11 & 23.23 & 17.55 & 392 \\
        \llamasmall & 14.82 & 79.83 & 25.00 & 5.88 & 31.68 & 9.92 & \textbf{1282} \\
        \rowcolor{MutedBlue} \llamasmall~w/ Elab. & 15.88 & \textbf{83.61} & 26.69 & 6.37 & \textbf{33.53} & 10.70 & 1253 \\
        \llamabig & 22.40 & 80.67 & 35.07 & 9.06 & 32.63 & 14.18 & 857 \\
        \rowcolor{MutedBlue} \llamabig~w/ Elab. & 16.89 & 68.07 & 27.07 & 7.08 & 28.54 & 11.35 & 959 \\
        \texttt{distilroberta-base}~($k=3$, finetuned) & 19.89 & 48.07 & 27.99 & 7.70 & 18.36 & 10.80 & - \\
        \texttt{distilroberta-base}~($k=5$, finetuned) & 18.32 & 46.43 & 26.16 & 8.42 & 20.97 & 11.97 & - \\
        \bottomrule
    \end{tabular}
    \caption{Macro-averaged precision, recall, and F1 scores of different models on detecting friction. Each friction instance contributes equally to the final metrics regardless of which conversation it appears in. \#Predictions refer to the total number of instances of conversational friction found by each model. For reference, annotators identified \textbf{238} instances in total. \gpt~\emph{without} Elaboration of technical terms (Sec \ref{sec:friction-annotation}) performed best across all models. Micro-averaged results are provided in the main text (Table~\ref{tab:results}).}
    \label{tab:macro-results}
\end{table*}

%% file: acl_latex.bbl
\begin{thebibliography}{42}
\providecommand{\natexlab}[1]{#1}

\bibitem[{Argyle et~al.(2023)Argyle, Bail, Busby, Gubler, Howe, Rytting, Sorensen, and Wingate}]{Argyle2023}
L.P. Argyle, C.A. Bail, E.C. Busby, J.R. Gubler, T.~Howe, C.~Rytting, T.~Sorensen, and D.~Wingate. 2023.
\newblock \href {https://doi.org/10.1073/pnas.2311627120} {Leveraging ai for democratic discourse: Chat interventions can improve online political conversations at scale}.
\newblock \emph{Proceedings of the National Academy of Sciences of the United States of America}, 120(41):e2311627120.

\bibitem[{Bara et~al.(2021)Bara, CH-Wang, and Chai}]{bara-etal-2021-mindcraft}
Cristian-Paul Bara, Sky CH-Wang, and Joyce Chai. 2021.
\newblock \href {https://doi.org/10.18653/v1/2021.emnlp-main.85} {{M}ind{C}raft: Theory of mind modeling for situated dialogue in collaborative tasks}.
\newblock In \emph{Proceedings of the 2021 Conference on Empirical Methods in Natural Language Processing}, pages 1112--1125, Online and Punta Cana, Dominican Republic. Association for Computational Linguistics.

\bibitem[{Bohus and Rudnicky(2008)}]{bohus2008sorry}
Dan Bohus and Alexander~I Rudnicky. 2008.
\newblock Sorry, i didn’t catch that!
\newblock In \emph{Recent trends in discourse and dialogue}, pages 123--154. Springer.

\bibitem[{Bonial et~al.(2022)Bonial, Hudson, Baker, Lukin, and Traum}]{bonialmaking}
Claire Bonial, Taylor Hudson, Anthony~L Baker, Stephanie~M Lukin, and David Traum. 2022.
\newblock Making sense of stop.
\newblock In \emph{AREA II Workshop 2022}.

\bibitem[{Canavan et~al.(1997)Canavan, Graff, and Zipperlen}]{callhome1997}
Alexandra Canavan, David Graff, and George Zipperlen. 1997.
\newblock Callhome american english speech ldc97s42.
\newblock Web Download.

\bibitem[{Castro(2017)}]{castro-2017-fast-krippendorff}
Santiago Castro. 2017.
\newblock Fast {K}rippendorff: Fast computation of {K}rippendorff's alpha agreement measure.
\newblock \url{https://github.com/pln-fing-udelar/fast-krippendorff}.

\bibitem[{Chandu et~al.(2021)Chandu, Bisk, and Black}]{chandu2021grounding}
Khyathi~Raghavi Chandu, Yonatan Bisk, and Alan~W Black. 2021.
\newblock Grounding'grounding'in nlp.
\newblock \emph{arXiv preprint arXiv:2106.02192}.

\bibitem[{Clark and Brennan(1991)}]{clark1991grounding}
Herbert~H. Clark and Susan~E. Brennan. 1991.
\newblock Grounding in communication.
\newblock In Lauren~B. Resnick, John~M. Levine, and Stephanie~D. Teasley, editors, \emph{Perspectives on Socially Shared Cognition}, pages 127--149. APA Books, Washington, DC.

\bibitem[{Clark and Wilkes-Gibbs(1986)}]{clark1986referring}
Herbert~H. Clark and Deanna Wilkes-Gibbs. 1986.
\newblock \href {https://doi.org/10.1016/0010-0277(86)90010-7} {Referring as a collaborative process}.
\newblock \emph{Cognition}, 22(1):1--39.

\bibitem[{Cohen(1960)}]{cohenkappa}
Jacob Cohen. 1960.
\newblock \href {https://doi.org/10.1177/001316446002000104} {A coefficient of agreement for nominal scales}.
\newblock \emph{Educational and Psychological Measurement}, 20(1):37--46.

\bibitem[{Cohen et~al.(2024)Cohen, Liu, Mooney, Tellex, and Watkins}]{cohen2024survey}
Vanya Cohen, Jason~Xinyu Liu, Raymond Mooney, Stefanie Tellex, and David Watkins. 2024.
\newblock A survey of robotic language grounding: Tradeoffs between symbols and embeddings.
\newblock \emph{arXiv preprint arXiv:2405.13245}.

\bibitem[{Geurts(2024)}]{sep-common-ground-pragmatics}
Bart Geurts. 2024.
\newblock {Common Ground in Pragmatics}.
\newblock In Edward~N. Zalta and Uri Nodelman, editors, \emph{The {Stanford} Encyclopedia of Philosophy}, {W}inter 2024 edition. Metaphysics Research Lab, Stanford University.

\bibitem[{Grosz and Sidner(1986)}]{grosz1986attention}
Barbara~J Grosz and Candace~L Sidner. 1986.
\newblock Attention, intentions, and the structure of discourse.
\newblock \emph{Computational linguistics}, 12(3):175--204.

\bibitem[{Harnad(1990)}]{harnad1990symbol}
Stevan Harnad. 1990.
\newblock The symbol grounding problem.
\newblock \emph{Physica D: Nonlinear Phenomena}, 42(1-3):335--346.

\bibitem[{Heeman and Allen(1994)}]{heeman1994detecting}
Peter Heeman and James Allen. 1994.
\newblock Detecting and correcting speech repairs.
\newblock \emph{arXiv preprint cmp-lg/9406006}.

\bibitem[{Hoyle et~al.(2023)Hoyle, Sarkar, Goel, and Resnik}]{hoyle-etal-2023-natural}
Alexander Hoyle, Rupak Sarkar, Pranav Goel, and Philip Resnik. 2023.
\newblock \href {https://doi.org/10.18653/v1/2023.emnlp-main.815} {Natural language decompositions of implicit content enable better text representations}.
\newblock In \emph{Proceedings of the 2023 Conference on Empirical Methods in Natural Language Processing}, pages 13188--13214, Singapore. Association for Computational Linguistics.

\bibitem[{Hu and Frank(2024)}]{hu2024auxiliarytaskdemandsmask}
Jennifer Hu and Michael~C. Frank. 2024.
\newblock \href {https://arxiv.org/abs/2404.02418} {Auxiliary task demands mask the capabilities of smaller language models}.
\newblock \emph{Preprint}, arXiv:2404.02418.

\bibitem[{Ilinykh et~al.(2019)Ilinykh, Zarrie{\ss}, and Schlangen}]{ilinykh2019meetup}
Nikolai Ilinykh, Sina Zarrie{\ss}, and David Schlangen. 2019.
\newblock Meetup! a corpus of joint activity dialogues in a visual environment.
\newblock \emph{arXiv preprint arXiv:1907.05084}.

\bibitem[{Khebour et~al.(2024)Khebour, Lai, Bradford, Zhu, Brutti, Tam, Tu, Ibarra, Blanchard, Krishnaswamy, and Pustejovsky}]{khebour-etal-2024-common}
Ibrahim~Khalil Khebour, Kenneth Lai, Mariah Bradford, Yifan Zhu, Richard~A. Brutti, Christopher Tam, Jingxuan Tu, Benjamin~A. Ibarra, Nathaniel Blanchard, Nikhil Krishnaswamy, and James Pustejovsky. 2024.
\newblock \href {https://aclanthology.org/2024.lrec-main.318} {Common ground tracking in multimodal dialogue}.
\newblock In \emph{Proceedings of the 2024 Joint International Conference on Computational Linguistics, Language Resources and Evaluation (LREC-COLING 2024)}, pages 3587--3602, Torino, Italia. ELRA and ICCL.

\bibitem[{Kummerfeld et~al.(2019)Kummerfeld, Gouravajhala, Peper, Athreya, Gunasekara, Ganhotra, Patel, Polymenakos, and Lasecki}]{kummerfeld-etal-2019-large}
Jonathan~K. Kummerfeld, Sai~R. Gouravajhala, Joseph~J. Peper, Vignesh Athreya, Chulaka Gunasekara, Jatin Ganhotra, Siva~Sankalp Patel, Lazaros~C Polymenakos, and Walter Lasecki. 2019.
\newblock \href {https://doi.org/10.18653/v1/P19-1374} {A large-scale corpus for conversation disentanglement}.
\newblock In \emph{Proceedings of the 57th Annual Meeting of the Association for Computational Linguistics}, pages 3846--3856, Florence, Italy. Association for Computational Linguistics.

\bibitem[{Levelt(1983)}]{levelt1983monitoring}
Willem~JM Levelt. 1983.
\newblock Monitoring and self-repair in speech.
\newblock \emph{Cognition}, 14(1):41--104.

\bibitem[{Lopes et~al.(2018)Lopes, Hemmingsson, and {\AA}strand}]{lopes-etal-2018-spot}
Jos{\'e} Lopes, Nils Hemmingsson, and Oliver {\AA}strand. 2018.
\newblock \href {https://aclanthology.org/L18-1305} {The spot the difference corpus: a multi-modal corpus of spontaneous task oriented spoken interactions}.
\newblock In \emph{Proceedings of the Eleventh International Conference on Language Resources and Evaluation ({LREC} 2018)}, Miyazaki, Japan. European Language Resources Association (ELRA).

\bibitem[{Lowe et~al.(2015)Lowe, Pow, Serban, and Pineau}]{lowe-etal-2015-ubuntu}
Ryan Lowe, Nissan Pow, Iulian Serban, and Joelle Pineau. 2015.
\newblock \href {https://doi.org/10.18653/v1/W15-4640} {The {U}buntu dialogue corpus: A large dataset for research in unstructured multi-turn dialogue systems}.
\newblock In \emph{Proceedings of the 16th Annual Meeting of the Special Interest Group on Discourse and Dialogue}, pages 285--294, Prague, Czech Republic. Association for Computational Linguistics.

\bibitem[{Markowska et~al.(2023)Markowska, Taghizadeh, Soubki, Mirroshandel, and Rambow}]{markowska-etal-2023-finding}
Magdalena Markowska, Mohammad Taghizadeh, Adil Soubki, Seyed Mirroshandel, and Owen Rambow. 2023.
\newblock \href {https://doi.org/10.18653/v1/2023.findings-emnlp.551} {Finding common ground: Annotating and predicting common ground in spoken conversations}.
\newblock In \emph{Findings of the Association for Computational Linguistics: EMNLP 2023}, pages 8221--8233, Singapore. Association for Computational Linguistics.

\bibitem[{Minaee et~al.(2024)Minaee, Mikolov, Nikzad, Chenaghlu, Socher, Amatriain, and Gao}]{minaee2024largelanguagemodelssurvey}
Shervin Minaee, Tomas Mikolov, Narjes Nikzad, Meysam Chenaghlu, Richard Socher, Xavier Amatriain, and Jianfeng Gao. 2024.
\newblock \href {https://arxiv.org/abs/2402.06196} {Large language models: A survey}.
\newblock \emph{Preprint}, arXiv:2402.06196.

\bibitem[{Mohapatra et~al.(2024)Mohapatra, Hassan, Romary, and Cassell}]{mohapatra-etal-2024-conversational}
Biswesh Mohapatra, Seemab Hassan, Laurent Romary, and Justine Cassell. 2024.
\newblock \href {https://aclanthology.org/2024.lrec-main.352} {Conversational grounding: Annotation and analysis of grounding acts and grounding units}.
\newblock In \emph{Proceedings of the 2024 Joint International Conference on Computational Linguistics, Language Resources and Evaluation (LREC-COLING 2024)}, pages 3967--3977, Torino, Italia. ELRA and ICCL.

\bibitem[{Narayan-Chen et~al.(2019)Narayan-Chen, Jayannavar, and Hockenmaier}]{narayan-chen-etal-2019-collaborative}
Anjali Narayan-Chen, Prashant Jayannavar, and Julia Hockenmaier. 2019.
\newblock \href {https://doi.org/10.18653/v1/P19-1537} {Collaborative dialogue in {M}inecraft}.
\newblock In \emph{Proceedings of the 57th Annual Meeting of the Association for Computational Linguistics}, pages 5405--5415, Florence, Italy. Association for Computational Linguistics.

\bibitem[{{OpenAI}(2024)}]{openai2024gpt4o}
{OpenAI}. 2024.
\newblock \href {https://openai.com/index/hello-gpt-4o/} {Hello gpt-4o: A new model for openai's future}.
\newblock Accessed: 2024-10-13.

\bibitem[{Ortmann(2022)}]{ortmann-2022-fine}
Katrin Ortmann. 2022.
\newblock \href {https://aclanthology.org/2022.lrec-1.150} {Fine-grained error analysis and fair evaluation of labeled spans}.
\newblock In \emph{Proceedings of the Thirteenth Language Resources and Evaluation Conference}, pages 1400--1407, Marseille, France. European Language Resources Association.

\bibitem[{Ramprasad et~al.(2024)Ramprasad, Ferracane, and Lipton}]{ramprasad2024analyzingllmbehaviordialogue}
Sanjana Ramprasad, Elisa Ferracane, and Zachary~C. Lipton. 2024.
\newblock \href {https://arxiv.org/abs/2406.03487} {Analyzing llm behavior in dialogue summarization: Unveiling circumstantial hallucination trends}.
\newblock \emph{Preprint}, arXiv:2406.03487.

\bibitem[{Sanh et~al.(2019)Sanh, Debut, Chaumond, and Wolf}]{Sanh2019DistilBERTAD}
Victor Sanh, Lysandre Debut, Julien Chaumond, and Thomas Wolf. 2019.
\newblock Distilbert, a distilled version of bert: smaller, faster, cheaper and lighter.
\newblock \emph{ArXiv}, abs/1910.01108.

\bibitem[{Shaikh et~al.(2024)Shaikh, Gligoric, Khetan, Gerstgrasser, Yang, and Jurafsky}]{shaikh-etal-2024-grounding}
Omar Shaikh, Kristina Gligoric, Ashna Khetan, Matthias Gerstgrasser, Diyi Yang, and Dan Jurafsky. 2024.
\newblock \href {https://doi.org/10.18653/v1/2024.naacl-long.348} {Grounding gaps in language model generations}.
\newblock In \emph{Proceedings of the 2024 Conference of the North American Chapter of the Association for Computational Linguistics: Human Language Technologies (Volume 1: Long Papers)}, pages 6279--6296, Mexico City, Mexico. Association for Computational Linguistics.

\bibitem[{Shaikh et~al.(2025)Shaikh, Mozannar, Bansal, Fourney, and Horvitz}]{shaikh2025navigatingriftshumanllmgrounding}
Omar Shaikh, Hussein Mozannar, Gagan Bansal, Adam Fourney, and Eric Horvitz. 2025.
\newblock \href {https://arxiv.org/abs/2503.13975} {Navigating rifts in human-llm grounding: Study and benchmark}.
\newblock \emph{Preprint}, arXiv:2503.13975.

\bibitem[{Stalnaker(1978)}]{stalnaker1978}
Robert Stalnaker. 1978.
\newblock Assertion.
\newblock \emph{Syntax and Semantics}, 9:315--332.

\bibitem[{Stalnaker(2002)}]{Stalnaker2002}
Robert Stalnaker. 2002.
\newblock \href {https://www.jstor.org/stable/25001871} {Common ground}.
\newblock \emph{Linguistics and Philosophy}, 25(5/6):701--721.
\newblock Accessed: 2019-02-15 07:19 UTC.

\bibitem[{Tan et~al.(2024)Tan, Westermann, Pottanigari, Šavelka, Meeùs, Godet, and Benyekhlef}]{tan2024robotsmiddleevaluatingllms}
Jinzhe Tan, Hannes Westermann, Nikhil~Reddy Pottanigari, Jaromír Šavelka, Sébastien Meeùs, Mia Godet, and Karim Benyekhlef. 2024.
\newblock \href {https://arxiv.org/abs/2410.07053} {Robots in the middle: Evaluating llms in dispute resolution}.
\newblock \emph{Preprint}, arXiv:2410.07053.

\bibitem[{Touvron et~al.(2023)Touvron, Lavril, Izacard, Martinet, Lachaux, Lacroix, Rozière, Goyal, Hambro, Azhar, Rodriguez, Joulin, Grave, and Lample}]{touvron2023llamaopenefficientfoundation}
Hugo Touvron, Thibaut Lavril, Gautier Izacard, Xavier Martinet, Marie-Anne Lachaux, Timothée Lacroix, Baptiste Rozière, Naman Goyal, Eric Hambro, Faisal Azhar, Aurelien Rodriguez, Armand Joulin, Edouard Grave, and Guillaume Lample. 2023.
\newblock \href {https://arxiv.org/abs/2302.13971} {Llama: Open and efficient foundation language models}.
\newblock \emph{Preprint}, arXiv:2302.13971.

\bibitem[{Traum and Allen(1992)}]{Traum1992}
David~R. Traum and James~F. Allen. 1992.
\newblock \href {https://doi.org/10.21437/ICSLP.1992-41} {A "speech acts" approach to grounding in conversation}.
\newblock In \emph{The Second International Conference on Spoken Language Processing, {ICSLP} 1992, Banff, Alberta, Canada, October 13-16, 1992}, pages 137--140. {ISCA}.

\bibitem[{Traum(1995)}]{Traum1995}
David~Rood Traum. 1995.
\newblock \emph{A Computational Theory of Grounding in Natural Language Conversation}.
\newblock Ph.D. thesis, University of the USA.
\newblock UMI Order No. GAX95-23171.

\bibitem[{Wang et~al.(2024)Wang, Xu, Li, Zhang, Liang, Tang, Yu, and Wen}]{wang2024largelanguagemodelseducation}
Shen Wang, Tianlong Xu, Hang Li, Chaoli Zhang, Joleen Liang, Jiliang Tang, Philip~S. Yu, and Qingsong Wen. 2024.
\newblock \href {https://arxiv.org/abs/2403.18105} {Large language models for education: A survey and outlook}.
\newblock \emph{Preprint}, arXiv:2403.18105.

\bibitem[{White et~al.(2017)White, Rastogi, Duh, and Van~Durme}]{white-etal-2017-inference}
Aaron~Steven White, Pushpendre Rastogi, Kevin Duh, and Benjamin Van~Durme. 2017.
\newblock \href {https://aclanthology.org/I17-1100} {Inference is everything: Recasting semantic resources into a unified evaluation framework}.
\newblock In \emph{Proceedings of the Eighth International Joint Conference on Natural Language Processing (Volume 1: Long Papers)}, pages 996--1005, Taipei, Taiwan. Asian Federation of Natural Language Processing.

\bibitem[{İnan et~al.(2025)İnan, Sicilia, Dey, Dongre, Srinivasan, Thomason, Tür, Hakkani-Tür, and Alikhani}]{i̇nan2025betterslowsorryintroducing}
Mert İnan, Anthony Sicilia, Suvodip Dey, Vardhan Dongre, Tejas Srinivasan, Jesse Thomason, Gökhan Tür, Dilek Hakkani-Tür, and Malihe Alikhani. 2025.
\newblock \href {https://arxiv.org/abs/2501.17348} {Better slow than sorry: Introducing positive friction for reliable dialogue systems}.
\newblock \emph{Preprint}, arXiv:2501.17348.

\end{thebibliography}
